# Feature importance analysis for patient management decisions

**Michal Valko**[a], **Milos Hauskrecht**[a]

[a] *Department of Computer Science, University of Pittsburgh, Pittsburgh, PA*

**Abstract**

*The objective of this paper is to understand what characteristics and features of clinical data influence physician's decision about ordering laboratory tests or prescribing medications the most. We conduct our analysis on data and decisions extracted from electronic health records of 4486 post-surgical cardiac patients. The summary statistics for 335 different lab order decisions and 407 medication decisions are reported. We show that in many cases, physician's lab-order and medication decisions are predicted well by simple patterns such as last value of a single test result, time since a certain lab test was ordered or time since certain procedure was executed.*

## Introduction

Advances in data collection and electronic health record technologies have led to the emergence of clinical datasets, where data instances consist of sequences of clinical findings, lab values, measurements, and medication actions [7]. Such multivariate time series data provide us with a complex temporal characterization of the patient case. Analyses of these clinical datasets can be extremely useful for building models supporting patient outcome prediction, early detection of adverse events, or clinical decision making [8].

The key challenge when analyzing the clinical datasets is the complexity of the multivariate time series and the number of possible temporal features (patterns) one may generate to characterize such data. Inevitably we ask what types of features are the most important to represent the patient case. Are patterns related to most recent patient history more important than the distant past? What features do the physicians base their decisions upon? Are *values* or *trends* more important? Do physicians tend to look into simple trends and simple time constrains or into more complex temporal characteristics between several clinical variables?

We study this problem by analyzing the importance of various temporal features for lab and medication order decisions. More specifically, we investigate what temporal characteristics of the patient state influence the physician's decision the most.

Our analyzes on a collection of 4486 post-surgical cardiac patient records show that a relatively simple temporal characterization of the patient state is often sufficient to predict well many lab order and medication decisions. Moreover, we identify which of those simple characteristics are the most valuable sources of information for such a prediction.

The paper is structured as follows. First, we introduce the post-surgical cardiac dataset and temporal features used in our analysis. After that, we analyze the data and present statistics reflecting how different features predict the lab order and medication decisions. Finally, we discuss the results and conclude.

## PCP Dataset

**Post-surgical cardiac patient (PCP) database** is a database of de–identified records for 4486 post–surgical cardiac patients treated at one of the University of Pittsburgh Medical Center (UPMC) teaching hospitals. The entries in the database were populated from data from the MARS system, which serves as an archive for much of the data collected at UPMC. The records for individual patients included discharge records, demographics, progress notes, all labs and tests (including standard and all special tests), two medication databases, micro-biology labs, EKG, radiology and special procedures reports, and a financial charges database. The data in PCP database were cleaned, cross-mapped, and are currently stored in a local MySQL database with protected access.

**Dataset used in the analysis.** To conduct our analysis, we used time-stamped data stored in the PCP database and converted them into a vector space representation of a patient state at discrete time points to get a collection of patient state examples. More specifically, each patient record in the PCP was used to build a sequence of patient state examples reflecting scenarios the physicians faced at 8:00am every 24 hours when managing the patient (Figure 1). Only the information available up to the segmentation points was considered in the vector space representation. Our 24-hour segmentation led to the total of 30,828 patient state examples.

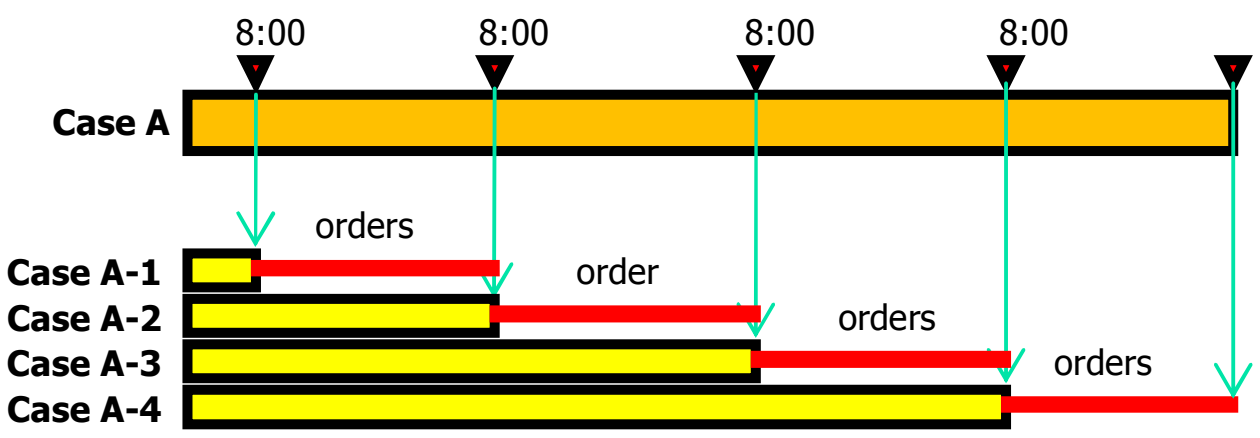

**Figure 1**. A segmentation of a patient case (Case A) to multiple patient state instances (A-1 to A-4) at 8:00am. Lab and medication orders for the following 24 hours are associated with each instance.

**Patient-management decisions.** In addition, every patient state example in the dataset that was generated by the above segmentation process was linked to lab order and medication decisions that were made for that patient within next 24 hours. Patient management decisions considered were:

- Lab order decisions with (true/false) values reflecting whether the lab was ordered within the next 24 hours or not
- Medication decisions with (true/false) values reflecting if the patient was given a medication within the next 24 hours or not.

A total of 335 lab order and 407 medication decision values were recorded and linked to every patient state example in the dataset.

### Features

To represent a patient state we have adopted a vector space representation that is convenient for machine learning approaches. In this representation a patient state is represented by a set of features characterizing the patient at a specific point in time and their corresponding feature values. Features represent and summarize the information in the medical record such as last blood glucose measurement, last glucose trend, or the time the patient is on heparin. These representations were also used in our experimental studies published in [1–3].

The features used in our experiment were generated from time series associated with different clinical variables, such as blood glucose measurement, platelet measurement, Amiodarone medication. The clinical variables used in this study were grouped into five categories:

1. Laboratory tests (LABs)
2. Medications (MEDs)
3. Visit features/demographics
4. Procedures
5. Heart support devices

We now briefly describe the features generated for clinical variables in each of these categories.

#### *Lab Features*

For the categorical labs, for example the ones with POS/NEG results we used the following features: Last value; second last value; first value; time since last order; whether the order pending; whether the value is known; and, whether the trend known. For the labs with continuous or ordinal values we used a richer set of features including features as difference between the last two values, slope of the last 2 values, and their percentage drop/increase. We used the same kind of features for the following pairs of lab values (last value, first value), (last value, nadir value), (last value, horizon value). Nadir and horizon value are the lab values with the smallest and the greatest value recorded up to that point. Figure 2 illustrates a subset of features generated for the labs with continuous values. The total number of features generated for such a lab is 40.

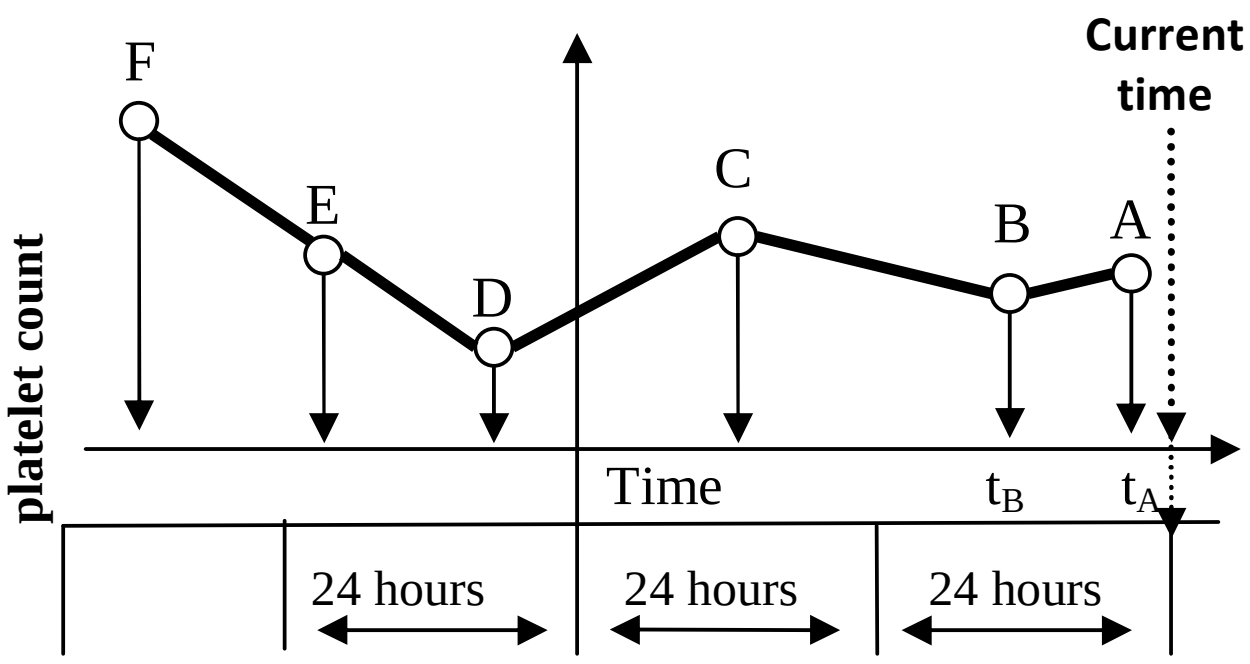

**Temporal features:**
Last value: A
Last value difference = B-A
Last percentage change = (B-A)/B
Last slope = (B-A) / ($t_B$-$t_A$)
Nadir = D
Nadir difference = A-D
Nadir percentage difference = (A-D)/D
Baseline = F
Drop from baseline = F-A
Percentage drop from baseline = (F-A)/F
24 hour average = (A+B)/2
…

**Figure 2**. A subset of temporal features generated for continuous valued lab tests.

#### *Medication Features*

For each medication we used four features: 1) indicator if the patient is currently on the medication 2) time since the patient was on that medication 3) time since the patient was put on that medication for the first time and, 4) time since last change in the status of patient taking the medication.

#### *Visit/Demographic Features*

We only have 3 features in this category: age, sex, and race. These are static and same for every time point we generate.

#### *Procedure Features*

Procedure features capture the information about procedures such as *Heart valve repair* that were performed either in OR or at the bedside. In our data we distinguish 36 different procedures that are performed on cardiac patients. We record four features per procedure: 1) time since the procedure was done last time 2) time since the procedure was done first time 3) whether the procedure was done in last 24 hours, and 4) whether the procedure was ever done to this patient.

***Heart Support Device Features***

Finally, we describe the status of 4 different heart support devices: Extracorporeal Membrane Oxygenation (ECMO), Balloon counter pulsation, Pacemaker, and other Heart Assist Device. For each of them, we record a single feature which describes whether the device is currently used to support patient's heart function.

Altogether, our dataset consists of 9,223 different features describing 30,828 patient states. As noted earlier, we use the patient state to evaluate our ability to predict 742 lab and medication order decisions.

## Methods

### Univariate AUC analysis

Our objective is to evaluate the significance of a feature for predicting either the lab order or medication order decision.

***Feature categories***

While one can always analyze predictive relations in between features and individual decisions, the aim of this paper is to understand what kinds of features influence lab and medication orders the most. Hence our analysis focuses on summary statistics across multiple labs and medication decisions, and across multiple feature categories. To conduct these analyses we grouped 9,223 different features into:

(1) Five categories corresponding to lab, medication, demographics, procedures, and heart support device features.

(2) Forty temporal feature categories, each representing the same temporal characteristic of the time series. For example the category 'Time since last LAB' subsumes 'Time since last Platelet count' and 'Time since last Glucose lab' features.

***Assessment metric***

We used the AUC score to assess the feature significance. AUC score is an area under the Receiver operating characteristic [4] which is used to measure the predictive strength of a feature. To assess the importance of each feature category we computed the number of times the feature in the category is the best AUC feature for predicting the decisions.

### Multivariate analysis

The limitation of the univariate analysis is the focus on an individual feature and it ability to predict the order decisions. In general, a better result may be often obtained by combining multiple features into a predictive model. To assess how helpful it is to use information from multiple features as opposed to just a single feature we have conducted a limited multivariate predictive analysis. Due the large amount of data, we used linear SVM classifier [5, 6]. In particular, for each order decision we trained such a classifier using 1) top 1; 2) top 3; and 3) top 30 features according to their AUC score computed on a subset of 2900 patients. After training the performance of the multivariate models was assessed by calculating their AUC on the remaining patient cases.

## Results

### Prediction of LAB orders

Figure 3 summarizes the most influential category for lab order decisions. The categories considered are labs, medication, procedure, demographics, and support devices features. Clearly, the most influential predictors for lab orders are features derived from lab and procedures data. Briefly, the best predictors for the next lab order are past labs. Intuitively, the lab order decision is typically driven by the existence of previous abnormal value of the same or other lab, and time since this lab has been measured. Procedure features are important as the type of surgical procedure and the time elapsed since the procedure may prompt close monitoring of certain organ functions and hence corresponding lab orders.

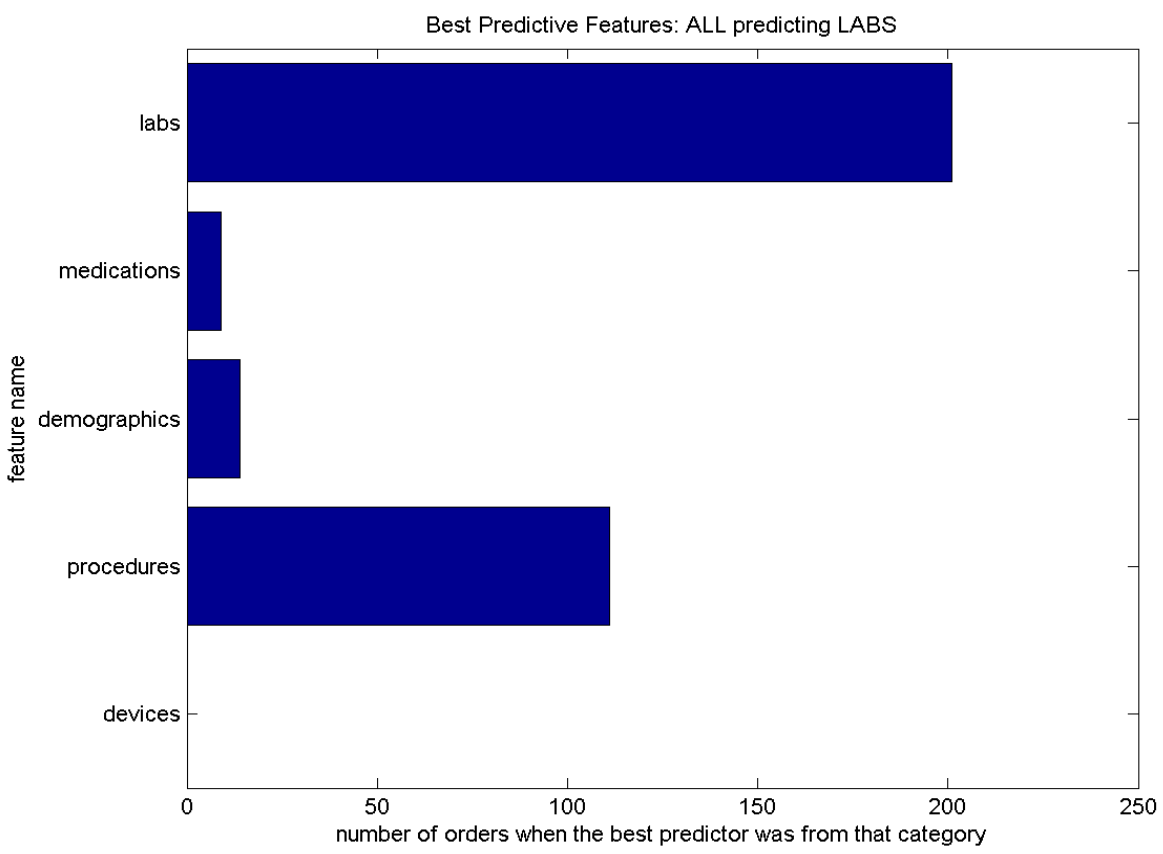

**Figure 3.** The importance of labs, meds, procedures, demographics and support device information for lab order decisions.

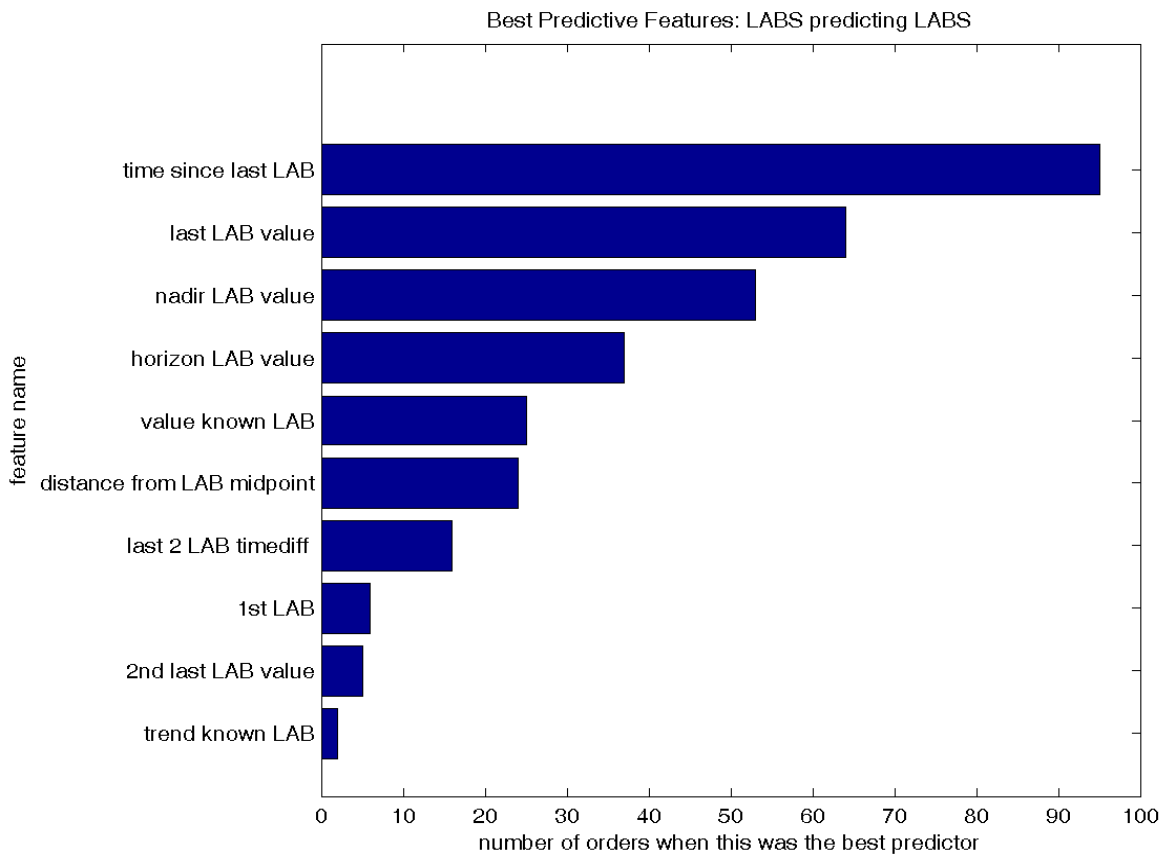

**Figure 4.** The most influential temporal lab feature categories for predicting lab order decisions with the same lab features.

The next histogram (Figure 4) shows the top 10 most important temporal lab features for predicting the orders of the same lab. The best feature categories were the times since the last lab order and lab results, in particular last value, nadir, and horizon values. The fact that the time since last measurement is the dominant feature is somewhat surprising but can be explained by the fact that many labs for our cohort of patients are done regularly and routinely, and the time since the last test was done is a good indicator of the upcoming lab order. Also surprising is a relatively low importance of trend features, in general absolute values of labs appear to be more significant for predicting the lab orders. This suggests value based prediction patterns are dominant for lab order decisions and trend information (if used at all) typically refines the pattern.

Figure 5 illustrates which temporal lab features predict the order of a different lab (i.e. not itself) the best. In this case, we observe that the last lab value is the most significant predictor of the order decision. This can be explained by the fact that an abnormal value of one test prompts the order of the other test. Also some lab tests are organized in panels and panels are typically ordered together creating this dependency.

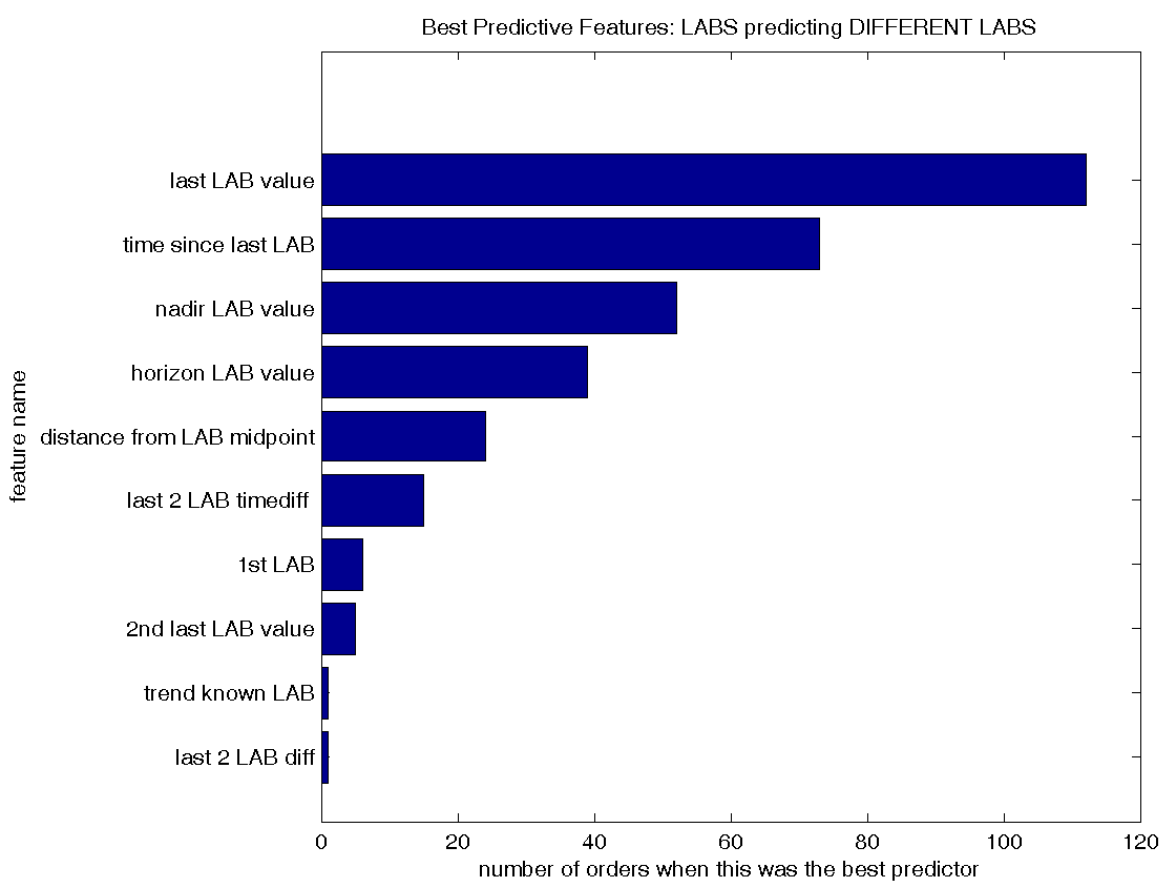

**Figure 5.** The most influential temporal feature categories for predicting lab order decisions with other lab features.

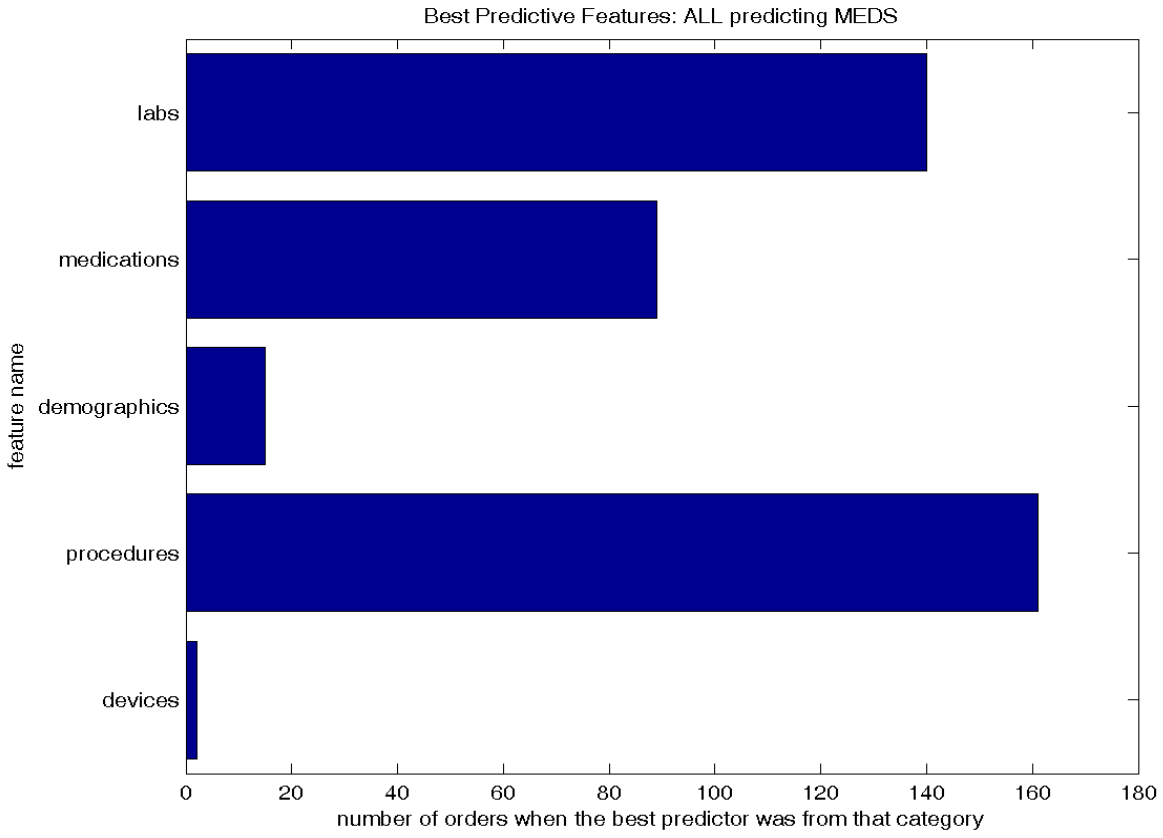

**Figure 6.** The importance of labs, meds, procedures, demographics, and support device information for medication commissions.

**Prediction of Medication orders**

Figure 6 shows the influence of labs, medication, demographics, procedures, and support device features for predicting medication orders. In particular, we are showing the prediction of a medication *commission*, provided that a patient was not on that medication before. We see that the list is dominated by procedure features as some medications tend to follow certain procedures.

Figure 7 breaks down the categories shown in Figure 6 and shows the top 10 most predictive features. For many medications the time since last procedure was the most predictive feature. For examples commissions of Papaverine could be most predicted (AUC= 95%) with the time that passed since the last Coronary Artery Bypass for that patient.

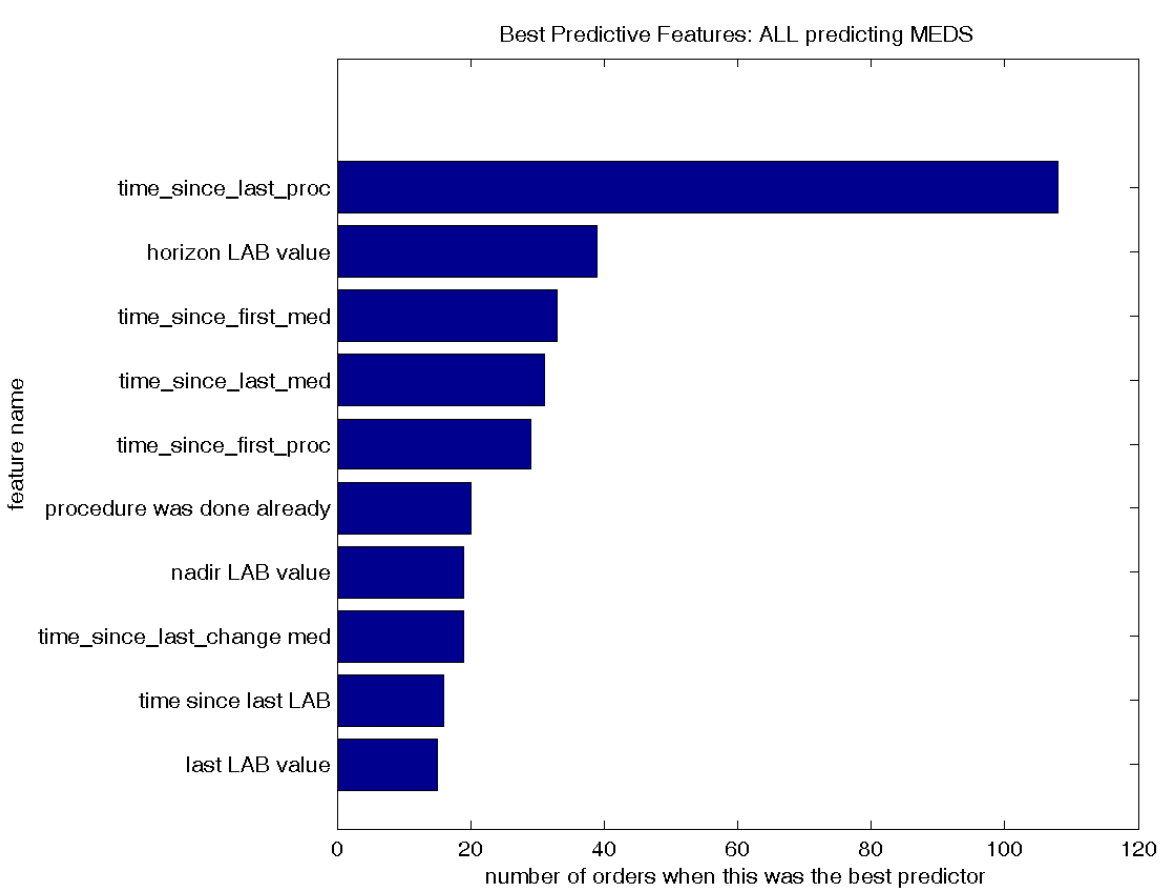

**Figure 7.** The most important temporal lab features for predicting medication commissions.

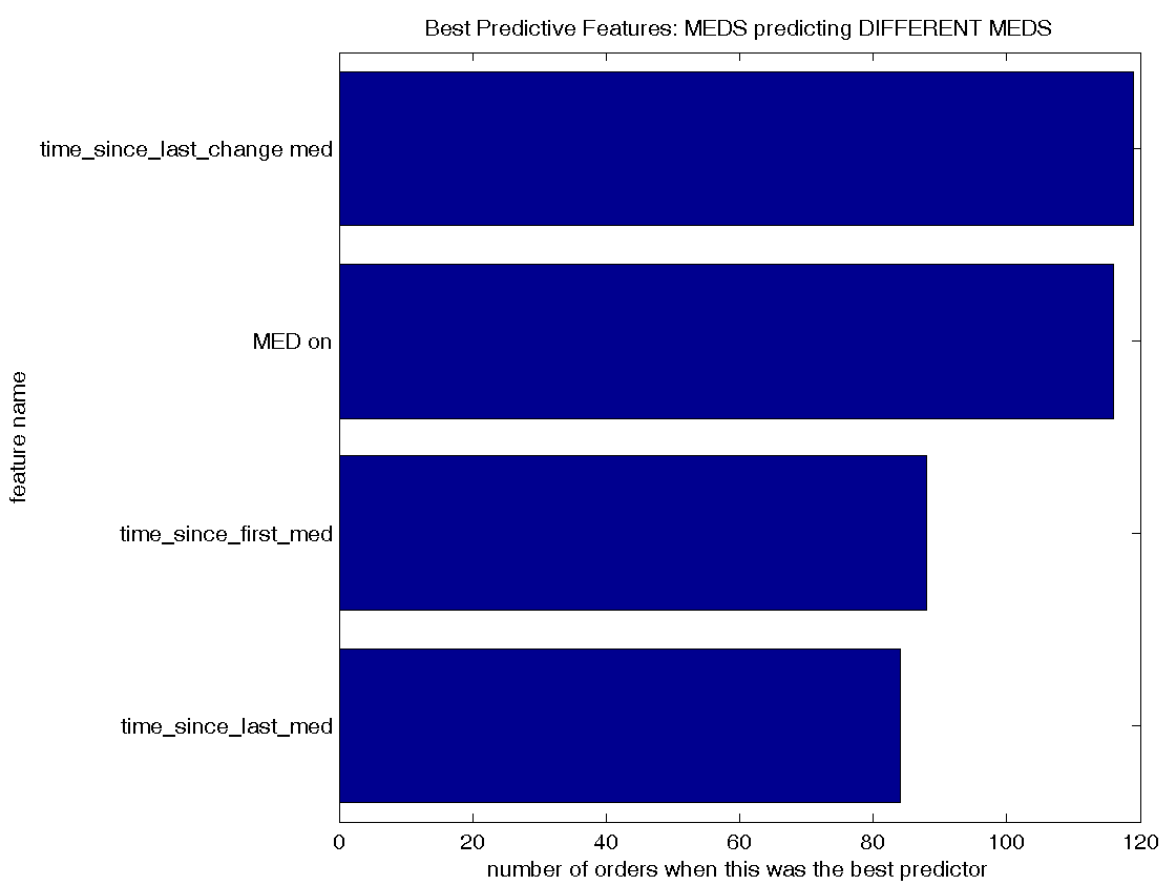

**Figure 8.** The importance of temporal features for predicting medication order decisions with other medication features.

Figure 8 shows the importance of medication features from different medications on the decision. One possible explanation for this dependency is that many medications are complementary and administered together, while other combinations are not used because of possible drug interaction. There-fore, a presence of one medication can often explain (predict) the presence or absence of another one.

### Multivariate prediction

The goal of this experiment is to explore the benefit of combining information from multiple sources or features. Tables 1 and 2 show AUC scores for prediction of top 10 highly predictive labs and medications from the linear SVM classifier. The classifier was trained on a subset of all features, in particular top 1, top 3 and top 30 best performing features in the training data. All results are reported on the independent test set.

We note that the feature selection for models with over 9000 feature candidates is an important and challenging problem and that the greedy selection applied in the experiment may not be the best one. However, this (somewhat limited) experiment indicates that a single feature is often a good predictor of the decision, suggesting simple patterns and their refinements may be able to capture well the prevalent lab order and medication order patterns.

*Table 1 AUC for linear SVM trained on top 1, 3 and 30 most predictive features, showing 10 highly predictive labs*

| | *top 1* | *top 3* | *top 30* |
|---|---|---|---|
| **GLU** | 87.46% | 85.97% | 87.73% |
| **TCPK** | 79.56% | 81.38% | 82.68% |
| **PPO2V** | 79.71% | 79.70% | 85.00% |
| **VANMCR** | 74.15% | 79.60% | 82.13% |
| **LD** | 79.22% | 79.14% | 82.69% |
| **HPA** | 83.54% | 83.03% | 84.97% |
| **RETAB2** | 75.83% | 75.67% | 81.89% |
| **TEGMA** | 75.91% | 85.36% | 84.76% |
| **TEGR** | 75.91% | 85.35% | 85.30% |
| **TEGALP** | 75.91% | 85.38% | 85.09% |

*Table 2 AUC for linear SVM trained on top 1, 3 and 30 most predictive features, showing 10 highly predictive medications*

| | *top 1* | *top 3* | *top 30* |
|---|---|---|---|
| **Nitroglycerin** | 71.48% | 79.50% | 80.53% |
| **Papaverine** | 82.34% | 83.83% | 89.19% |
| **Ioversol** | 87.65% | 88.58% | 89.50% |
| **Aminocaproic** | 79.49% | 79.39% | 86.83% |
| **Aprotinin** | 80.67% | 78.65% | 85.99% |
| **Thiopental** | 76.40% | 75.62% | 85.30% |
| **Eptifibatide** | 87.22% | 89.13% | 89.75% |
| **Darbepoetin** | 86.22% | 90.33% | 88.85% |
| **Iodixanol** | 69.62% | 87.99% | 87.26% |
| **Vitamin K** | 85.87% | 86.08% | 86.61% |

## Conclusions

Our univariate analyses of relations in between patient states and patient-management decisions revealed that the lab and medication order decisions are often driven by simple predictive patterns that involve more recent set of values, or times since the occurrence of some event (e.g. procedures, or previous lab/medication orders). Our (limited) analyses of more complex multivariate models suggest that lab and medication order decisions are likely based on only few clinical variables and their characteristics (features). In the future, we plan to further expand this study by analyzing feature dependencies and by developing more advanced feature selection for building multivariate predictive models.

### Acknowledgments

The research presented in this paper was funded by grants R21-LM009102-01A1 and 1R01LM010019-01A1. Its content is solely the responsibility of the authors and do not necessarily represent the official views of the NIH.

Milos Hauskrecht, Computer Science Department, University of Pittsburgh, 210 S Bouquet St, 5329 Sennott Sq, , Pittsburgh, PA, USA, milos@cs.pitt.edu